\documentclass[letterpaper, 10 pt, conference]{ieeeconf}  

\IEEEoverridecommandlockouts                              

\usepackage{graphicx} 
\usepackage{epsfig} 
\usepackage{mathptmx} 
\usepackage{times} 
\usepackage{amsmath} 
\usepackage{amssymb}  
\usepackage{tabularx}
\usepackage{comment}
\usepackage{hyperref}
\usepackage{cite}
\usepackage{booktabs}
\usepackage{multirow}
\usepackage{censor}

\title{\LARGE \bf
OCELOT: Odometry and Contact Estimation for Legged Robots
}

\author{Emre Girgin$^{1}$ and Cagri Kilic$^{1}$
\thanks{$^{1}$Department of Aerospace Engineering, Embry-Riddle Aeronautical University, Daytona Beach, FL 32114, USA.
        {\tt\small girgine@my.erau.edu}}%
 }

\begin{document}

\newcommand{\vect}[1]{\bm{#1}}
\newcommand{\matr}[1]{\bm{#1}}

\maketitle
\thispagestyle{empty}
\pagestyle{empty}

\begin{abstract}

One of the significant challenges in legged robotics is achieving accurate odometry using only onboard proprioceptive sensors. In this study, we present a complete leg odometry pipeline based on an Error-State EKF (ESEKF) that relies exclusively on proprioceptive data: a body fixed IMU, joint encoders, and force sensors, where filter's state is corrected by feet determined to be in a stationary stance. The core of our contribution is fused contact detection and an uncertainty quantification module designed to explicitly identify and reject slippage. This module runs two detectors in parallel for each foot, 1) a debounced, force-based Gaussian Mixture Model (GMM) guided Finite State Machine (FSM) to confirm physical contact, and 2) a kinematic-based Generalized Likelihood Ratio Test (GLRT) on the estimated velocity of the foot. The continuous quality scores from both estimators are fused to detect if the foot is both physically loaded and kinematically stationary and served as an uncertainty signal for each contact. To validate our approach, we collected a multi-modal dataset of 29 sequences spanning diverse indoor and outdoor terrains (e.g., concrete, grass, pebble, and rock) total of 2.4 km long. We benchmarked our approach against both proprioceptive and exteroceptive methods. The results demonstrate our method's efficacy in providing accurate odometry estimates, robustly handling slippage-prone environments. We also share our code and real-time ROS2 package as open-source. \footnote{\href{https://github.com/srge-erau/leg_odometry_uncertainty}{Github}}
\end{abstract}

\vspace{-0.2cm}

\section{INTRODUCTION}

Legged robots are receiving significant research interest due to their better locomotion capabilities in unstructured and uneven terrains, environments that pose considerable challenges for conventional wheeled robots \cite{semini2020legged, chung2023into}. 
On the other hand, the absence of wheels necessitates a more complex locomotion strategy, characterized by intermittent contact dynamics as the end-effectors (feet) cyclically make and break contact with the ground \cite{sh-ch12-proprio}. Tracking the position and velocity of the robot's main body, or \textit{Leg Odometry} \cite{roston1991dead}, is achieved by exploiting the robot's kinematic and dynamic models \cite{bloesch2013state}. 
Consequently, the fundamental proprioceptive sensor suite for a legged robot typically comprises an IMU and joint encoders to measure angular positions and velocities; this contrasts with the IMU and wheel encoder pairing used on wheeled platforms. While these sensors are standard, the suite is frequently augmented with contact/force sensors to provide explicit information regarding foot-ground interaction.

\begin{figure}[t]
    \centering
    \includegraphics[width=1.0\linewidth,trim={0 14cm 0 0cm},  
    clip]{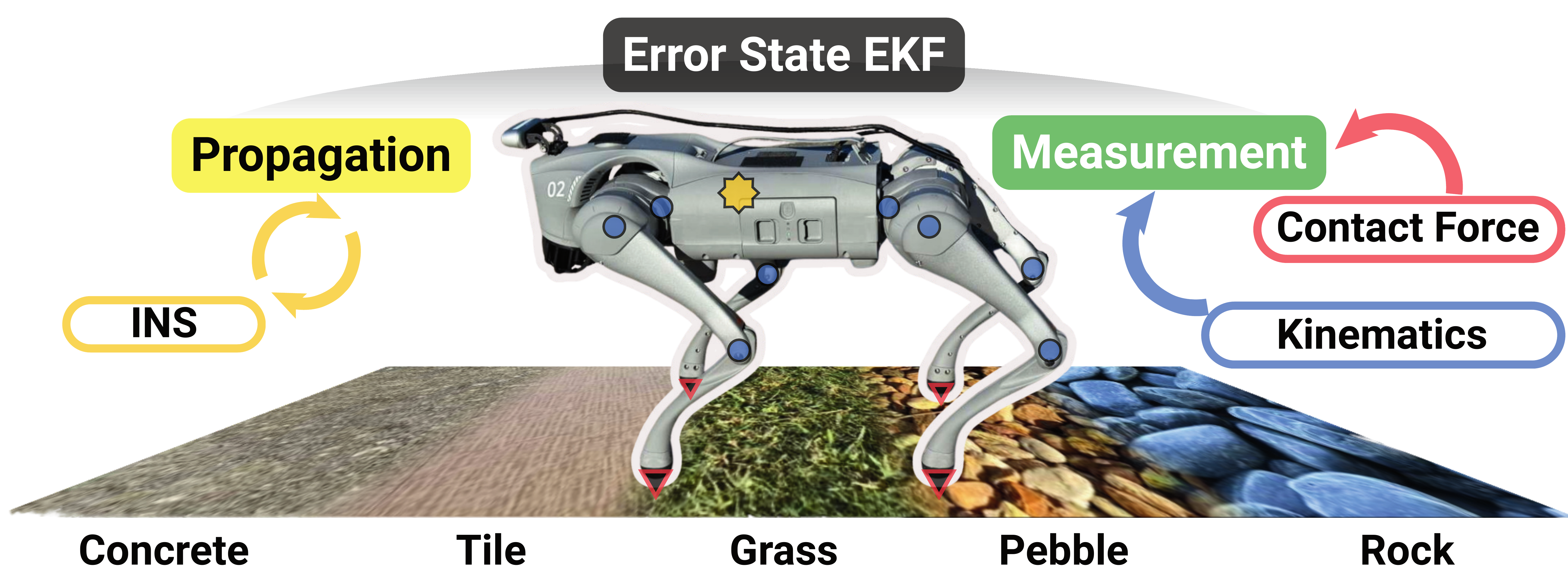}
    \caption{Our contact estimation method provides robustness against various terrain, including rough granular medium like rock and pebble leading to slippage.}
    \label{fig:go2_main}
    \vspace{-0.7cm}
\end{figure}

The dominant framework in leg odometry is to propagate the robot's state via an IMU-driven Inertial Navigation System (INS) and correct drift through Zero Velocity Updates (ZUPTs) \cite{skog2010zero} when a foot is determined to be in stance. However, accurately applying this correction requires knowing exactly when a leg is stationary, making contact state estimation a critical sub-problem. Existing solutions, reviewed in Section~\ref{sec:related_work}, range from force-based thresholding to leg-mounted IMUs and learning-based classifiers.

In this study, we adopt an ESEKF architecture that uses only the onboard proprioceptive sensor suite of IMU, joint encoders, and foot force sensors. The core of our contribution lies in fused contact detection and uncertainty quantification module. This module runs two detectors in parallel: a force-based FSM which is adaptively governed by a GMM for direct, physical contact, and a kinematic GLRT on estimated foot velocity. By fusing the continuous quality scores from both detectors, our method robustly identifies stance phases which are physically loaded and kinematically stationary. This fused score is then used to dynamically compute an adaptive measurement covariance for each stance based on contact quality. This allows the ESEKF to intelligently weight each measurement regarding its reliability: a high-quality, non-slip contact results in a low measurement covariance and a strong correction, while a low-quality event like a slip results in a high measurement covariance, causing the filter to automatically down-weight its influence. Finally, an innovation based gating step validates each accepted measurement, rejecting any remaining statistical outliers before the final correction.
\begin{figure*}[thpb]
  \centering
  \includegraphics[
    scale=0.18,
    trim={0 19.5cm 0 0cm},  
    clip
  ]{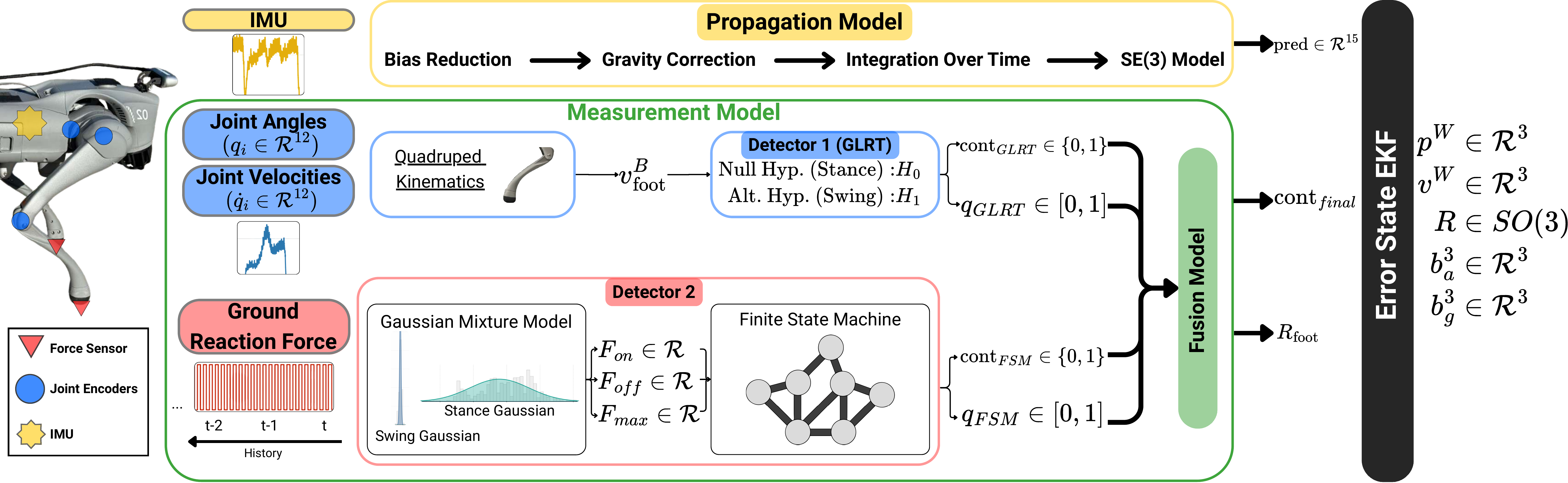}
  \caption{Overall architecture of the proposed method. Our study utilizes IMU for state propagation and corrects the state based on the contact event for each leg separately. Detecting contact and quantifying the uncertainty is conducted by two distinct detectors and their predictions are fused as the measurement model.}
  \label{fig:architecture}
  \vspace{-0.7cm}
\end{figure*}
We benchmarked our method against both proprioceptive-only methods and exteroceptive approaches and demonstrated our method's effectiveness. To summarize, our contributions in this paper are:
\begin{itemize}

    \item A novel contact detection and adaptive uncertainty framework that combines a force-based GMM-FSM and a body IMU derived GLRT contact stationarity detector, eliminating the need for supplementary foot-mounted sensors, to generate a continuous contact quality score, which is translated into a dynamic measurement covariance $\mathbf{R}_i$.
    \item An open-source ESEKF-based state estimator for quadruped leg odometry using only onboard proprioceptive sensors (fixed-body IMU, joint encoders, and force sensors).
    \item A new publicly available legged odometry dataset for quadrupeds, including 29 indoor and outdoor sequences collected on diverse terrain types (concrete, tile, grass, pebble and rock) for total of 2.4 km path.
\end{itemize}

\vspace{-0.3cm}

\section{RELATED WORK}
\label{sec:related_work}

Proprioceptive leg odometry has been extensively studied using EKFs \cite{bloesch2013state, bloesch2013stateVelocity}, Invariant EKF (InEKF) \cite{bonnabel2007left, barrau2016invariant} frameworks \cite{hartley2018legged, hartley2018contact, teng2021legged, Hartley-RSS-18,santana2024proprioceptive, kim2025adaptive}, smoothing-based approaches \cite{kim2021legged, yoon2023invariant}, and learning-based alternatives \cite{buchanan2022learning, wasserman2024legolas}. All these various frameworks mainly adopt IMU based propagation and differ by the contact detection method they obtained. Moreover, all these studies highlight a fundamental limitation: the unobservability of absolute position and heading when relying solely on body-fixed IMUs, joint encoders, and force sensors \cite{bloesch2013state, bloesch2013stateVelocity, teng2021legged, hartley2018legged} making proprioceptive leg odometry a challenging problem.

\subsection{Contact Estimation}

One major category of contact estimation methods relies directly on Ground Reaction Force (GRF) data. Fallon et al. \cite{fallon2014drift} constructed an FSM based on torque/force measurements to provide debounced contact states, relying on fixed state transitions to determine contact events. Rotella et al. \cite{rotella2018unsupervised} applied weighted K-means clustering to GRF data to automatically group measurements into distinct contact and swing phases. Conversely, this method assumes the GRF signal space is separable into simple spherical clusters. A second category of methods infers contact without direct force measurements, often relying on robot dynamics or kinematics. Bloesch et al. \cite{bloesch2013state, bloesch2013stateVelocity} adopts simple thresholding on estimated GRF with static measurement covariance. Camurri et al. \cite{camurri2017probabilistic} estimated the GRF by formulating a dynamic model that integrated joint angles, motor efforts, and torques generated by external forces using fixed thresholding on the GRF and does not explicitly address foot slippage. Jenelten et al. \cite{jenelten2019dynamic} proposed a kinematic approach, fitting the vertical component of the estimated foot velocity to an exponential distribution and interpreting peaks in its derivative as touchdown events. While this method rapidly detects slippage, it struggles to consistently classify the complete stance phase. Hartley et al. \cite{Hartley-RSS-18} estimated the GRF from joint torques and applied logistic regression to adaptively compute contact probability, similar to ours. However, the associated covariance is primarily governed by $\Delta$GRF, making it sensitive to the initial touchdown event, rather than reflecting the overall quality of the complete stance phase.

Another approach involves augmenting the robot with additional sensors on the legs. Yang et al. \cite{yang2023multi}  and equipped each leg with an additional IMU to capture the end-effector's inertial motion, detecting contact by observing the discrepancy between the IMU's measurement and the kinematically-derived velocity. Similarly, the DogLegs framework \cite{wu2025doglegs} utilized leg-mounted IMUs to apply a GLRT adapted from pedestrian tracking \cite{skog2010zero}. In contrast, the proposed method eliminates the need for supplementary hardware. We rely solely on the standard body-fixed IMU for state propagation and utilize joint angles exclusively for our GLRT-based contact detector. Finally, learning-based approaches have recently been introduced. Lin et al. \cite{lin2021legged} developed a 1D CNN-based kinematics estimator for quadruped robots. Youm et al. \cite{youm2025legged} proposed a Gated Recurrent Unit (GRU) based neural network that processes the robot's target velocity commands and previous state to estimate both contact probability and body velocity. These learned outputs are then fused within an InEKF estimator. Although the accuracy of these learning-based classifiers is promising, the black-box nature of neural networks yields uncalibrated uncertainty \cite{sensoy2018evidential}, a critical limitation further evaluated in our experiments.
\vspace{-0.3cm}
\subsection{Quadruped Datasets}
To rigorously evaluate the proposed methodology, we introduce a comprehensive, custom-collected dataset. Existing quadrupedal datasets exhibit specific limitations that restrict their utility for multi-modal odometry evaluation. For instance, the Legkilo \cite{10631676},  Co-RaL \cite{jung2024co} and UMich Deep Contact Estimation \cite{lin2021legged} datasets contain solely proprioceptive measurements lacking RGB video streams and are predominantly confined to rigid surfaces, with the exception of grass, precluding benchmark comparisons against exteroceptive methods. The Cerberus dataset \cite{yang2023cerberus} omits contact data and lacks exposure to complex, challenging terrains.

In contrast, the proposed dataset spans a diverse spectrum of environments, incorporating rigid surfaces, deformable grounds, and challenging granular terrains. We provide synchronized proprioceptive data alongside RGB video streams, supplemented by point cloud data for indoor sequences. Finally, comprising a cumulative trajectory of approximately 2.4 km, our dataset offers longer continuous evaluation duration compared to the aforementioned alternatives.

\vspace{-0.1cm}

\section{METHODOLOGY}

Our framework is based on an ESEKF where update step is driven by our novel contact estimator as summarized in Figure \ref{fig:architecture}.
\vspace{-0.2cm}
\subsection{Preliminaries: Error-State Kalman Filter Design}

\subsubsection{Kinematics}
For a single leg $i$, the forward kinematics $\mathbf{fk}_i$ and Jacobian $\mathbf{J}_{v,i}$ map the joint angles $\mathbf{q}_i \in \mathbb{R}^N$ and velocities $\dot{\mathbf{q}}_i$ to the foot's position $\mathbf{p}_i^B$ and linear velocity $\mathbf{v}_i^B$ in the body frame $\{B\}$:
\begin{equation}
    \mathbf{p}_i^B = \mathbf{fk}_i(\mathbf{q}_i), \quad \mathbf{v}_i^B = \mathbf{J}_{v,i}(\mathbf{q}_i) \dot{\mathbf{q}}_i .
\end{equation}

\noindent We employ an ESEKF to fuse proprioceptive sensor data, which separates the state into a non-linear nominal state $\mathbf{x}$ and a linear error state $\delta\mathbf{x}$.

\subsubsection{State Definition}
The nominal state $\mathbf{x}$ defines the robot's pose and IMU biases: position $\mathbf{p}^W \in \mathbb{R}^3$ and velocity $\mathbf{v}^W \in \mathbb{R}^3$ of the body frame $\{B\}$ in the world frame $\{W\}$, rotation matrix $\mathbf{R} \in SO(3)$ transforming from $\{B\}$ to $\{W\}$, additive accelerometer ($\mathbf{b}_a \in \mathbb{R}^3$) and gyroscope ($\mathbf{b}_g \in \mathbb{R}^3$) biases in $\{B\}$. The error state $\delta\mathbf{x} = [\delta\mathbf{p}^T, \delta\mathbf{v}^T, \delta\boldsymbol{\theta}^T, \delta\mathbf{b}_a^T, \delta\mathbf{b}_g^T]^T \in \mathbb{R}^{15}$ is a small, zero-mean Gaussian perturbation composed with the nominal state via addition for vectors and the exponential map $\text{Exp}(\cdot): \mathbb{R}^3 \to SO(3)$ \cite{sola2018micro} for rotation.

\subsubsection{Prediction Step (Propagation)}
The nominal state is propagated by integrating bias-corrected IMU measurements ($\mathbf{a}^B = \mathbf{a}_{\text{raw}} - \mathbf{b}_a$, $\boldsymbol{\omega}^B = \boldsymbol{\omega}_{\text{raw}} - \mathbf{b}_g$) via Euler integration.

\paragraph{Error Covariance Propagation}
The error covariance $\mathbf{P}$ is propagated using the linearized error dynamics:
\begin{equation}
    \mathbf{P}_k = \mathbf{F}_d \mathbf{P}_{k-1} \mathbf{F}_d^T + \mathbf{Q}_d
\end{equation}

\noindent The discretized state transition matrix $\mathbf{F}_d \approx \mathbf{I} + \mathbf{F}_c \Delta t$ and process noise $\mathbf{Q}_d = \mathbf{G} \mathbf{Q}_c \mathbf{G}^T \Delta t$ are derived from the continuous error dynamics $\mathbf{F}_c$ and noise input $\mathbf{G}$ (where $[\cdot]_\times$ is the skew-symmetric operator and $\mathbf{Q}_c$ is a $12\times12$ noise matrix):
\begin{equation}
    \setlength{\arraycolsep}{1.5pt} 
    \mathbf{F}_c =
    \begin{bmatrix}
        \mathbf{0} & \mathbf{I} & \mathbf{0} & \mathbf{0} & \mathbf{0} \\
        \mathbf{0} & \mathbf{0} & - \mathbf{R} [\mathbf{a}^B]_\times & - \mathbf{R} & \mathbf{0} \\
        \mathbf{0} & \mathbf{0} & -[\boldsymbol{\omega}^B]_\times & \mathbf{0} & -\mathbf{I} \\
        \mathbf{0} & \mathbf{0} & \mathbf{0} & \mathbf{0} & \mathbf{0} \\
        \mathbf{0} & \mathbf{0} & \mathbf{0} & \mathbf{0} & \mathbf{0}
    \end{bmatrix}, \quad
    \mathbf{G} =
    \begin{bmatrix}
        \mathbf{0} & \mathbf{0} & \mathbf{0} & \mathbf{0} \\
        -\mathbf{R} & \mathbf{0} & \mathbf{0} & \mathbf{0} \\
        \mathbf{0} & -\mathbf{I} & \mathbf{0} & \mathbf{0} \\
        \mathbf{0} & \mathbf{0} & \mathbf{I} & \mathbf{0} \\
        \mathbf{0} & \mathbf{0} & \mathbf{0} & \mathbf{I}
    \end{bmatrix}
\end{equation}

\subsubsection{Correction Step (Zero-Velocity Update)}
The core assumption of leg odometry is that a foot in contact with the ground (in "stance") is stationary relative to the world frame $\{W\}$, $\mathbf{v}_{\text{foot}, i}^W = \mathbf{0}$.

\paragraph{Measurement Model}

The measurement model $h(\mathbf{x}_k)$ estimates the stance foot's world-frame velocity by combining the body velocity $\mathbf{v}_k^W$ with the relative foot velocity $\mathbf{v}_{\text{rel},i}^B$ (derived from angular velocity $\boldsymbol{\omega}^B$, foot position $\mathbf{p}_i^B$, and joint kinematics):
\begin{equation} \label{eqn:measurement_model}
    h(\mathbf{x}_k) = \mathbf{v}_k^W + \mathbf{R}_k \mathbf{v}_{\text{rel},i}^B, \quad
    \mathbf{v}_{\text{rel},i}^B = \boldsymbol{\omega}^B \times \mathbf{p}_i^B + \mathbf{J}_{v,i}(\mathbf{q}_i)\dot{\mathbf{q}}_i
\end{equation}

\paragraph{Measurement Jacobian}
The measurement Jacobian $\mathbf{H}_k = \partial h / \partial \delta\mathbf{x}$ for foot $i$ is a $3 \times 15$ matrix:
\begin{equation}
    \mathbf{H}_k =
    \begin{bmatrix}
        \mathbf{0}_{3\times3} & \mathbf{I}_{3\times3} & - \mathbf{R}_k [\mathbf{v}_{\text{rel},i}^B]_\times & \mathbf{0}_{3\times3} & \mathbf{R}_k [\mathbf{p}_i^B]_\times
    \end{bmatrix}
\end{equation}

\paragraph{Innovation and Gating}

To reject outliers (e.g., slips), the innovation $\mathbf{\nu}_k = -h(\mathbf{x}_k)$  ($\mathbf{y}_k = \mathbf{0}$)  is subjected to a $\chi^2$ gating test \cite{bloesch2013stateVelocity} using the covariance $\mathbf{S}_k = \mathbf{H}_k \mathbf{P}_k \mathbf{H}_k^T + \mathbf{R}_k$:
\begin{equation}
    \mathbf{\nu}_k = \mathbf{y}_k - h(\mathbf{x}_k), \quad \text{Reject if: } \mathbf{\nu}_k^T \mathbf{S}_k^{-1} \mathbf{\nu}_k > \gamma_{95}
\end{equation}

\paragraph{State and Covariance Update}

If accepted, the filter computes the gain $\mathbf{K}_k$ and error $\delta\mathbf{x}^+$, updates the covariance $\mathbf{P}_k^+$ (Joseph form), and injects errors into the nominal state (additive for vectors, manifold for $\mathbf{R}$):
\begin{equation}
    \begin{aligned}
        \mathbf{K}_k &= \mathbf{P}_k \mathbf{H}_k^T \mathbf{S}_k^{-1}, \quad \delta\mathbf{x}^+ = \mathbf{K}_k \mathbf{\nu}_k \\
        \mathbf{P}_k^+ &= (\mathbf{I} - \mathbf{K}_k \mathbf{H}_k) \mathbf{P}_k (\mathbf{I} - \mathbf{K}_k \mathbf{H}_k)^T + \mathbf{K}_k \mathbf{R}_{k} \mathbf{K}_k^T \\
        \mathbf{R}^+ &= \text{Exp}(\delta\boldsymbol{\theta}^+) \mathbf{R}, \quad \mathbf{z}^+ = \mathbf{z} + \delta\mathbf{z}^+ \;\; \forall \mathbf{z} \in \{\mathbf{p}, \mathbf{v}, \mathbf{b}_a, \mathbf{b}_g\}
    \end{aligned}
\end{equation}

\vspace{-0.2cm}
\subsection{Contact Detection}
\label{sec:contact_detection}

We utilize a fused contact detection method to achieve a robust stance quality estimation to the complementary failure modes of its components. The first component, a force based adaptive GMM guided FSM, provides a \textit{direct, non-recursive} measurement of physical contact but is blind to foot slippage. The second detector, a velocity based GLRT provides a \textit{kinematic} measurement of stationarity but can be susceptible to sensor noise. By combining both signals, we create a strict test that validates a measurement update only when a foot is \textbf{both} physically loaded \textbf{and} kinematically stationary. This fusion provides robust rejection of foot slip events, which are critical outlier cases (high force, non-zero velocity) that would otherwise corrupt the ESEKF state.

\subsubsection{Detector 1: GMM Guided FSM (GMM-FSM)}

\begin{figure}[t]
  \centering
  \includegraphics[scale=0.35]{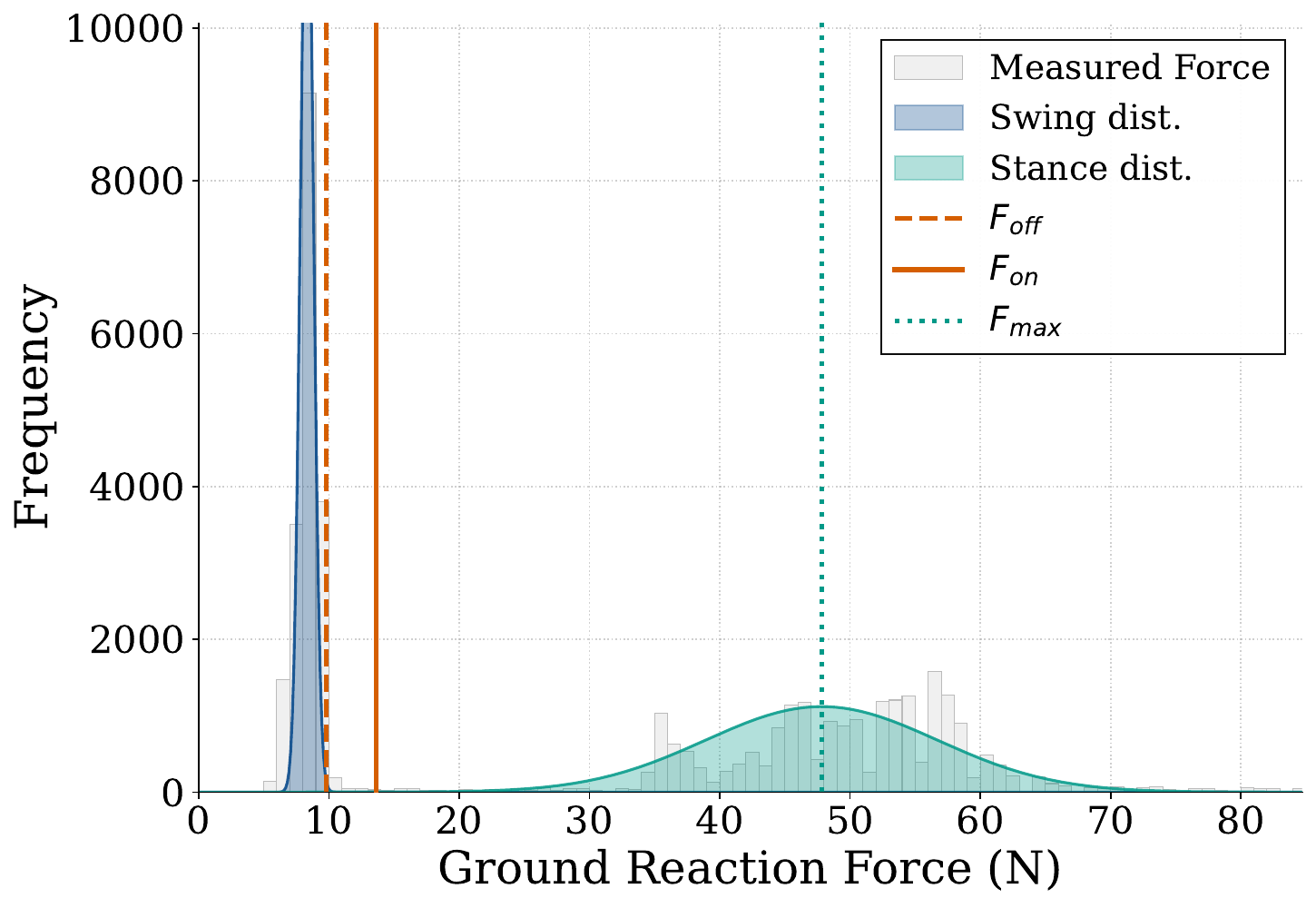}
  \caption{Gaussian Mixture Model fitted force histogram for each leg.}
  \label{fig:gmm}
  \vspace{-0.6cm}
\end{figure}

To provide robust, debounced contact estimation, we implement an FSM governing the discrete transitions of each foot between swing and stance based on the current state and the observed ground reaction force. The dynamically generated state transition thresholds, $F_{\text{on}}$ and $F_{\text{off}}$, determine the touchdown and liftoff events, respectively. To avoid the manual tuning of the state transition variables, we automatically determine these thresholds. The physical reality of a walking robot dictates that the force magnitude distribution, $F_{\text{mag}}$, is inherently bimodal, exhibiting a low-force \textbf{SWING} peak and a high-force \textbf{STANCE} peak. We fit a two-component GMM to a sliding window of recent historical data, comprising the last 10,000 samples (20 s at 500 Hz) updated at every new 2,500 samples (5 s), for each leg to autonomously identify the swing $\mathcal{N}(\mu_{s}, \sigma_{s})$ and stance $\mathcal{N}(\mu_{t}, \sigma_{t})$ distributions. 
The operational thresholds (visualized in Figure \ref{fig:gmm}) are then derived directly from these statistical parameters: $F_{\text{max}} = \mu_{t}$ (mean of the stance distribution), $F_{\text{off}} = \mu_{s} + 3\sigma_{s}$ (robust boundary encompassing 99.7\% of swing data noise), and $F_{\text{on}} = F_{\text{off}} + 0.1 \times (F_{\text{max}} - F_{\text{off}})$ (hysteresis margin).
By continuously fitting the GMM to only this recent data, the thresholds dynamically adapt to varying surface characteristics. This adaptive formulation renders the contact detection method robust across different terrains without any tuning, as demonstrated in Section \ref{sec:experiments}. For the failure cases disturbing bimodal distribution such as standing long durations, we switch to the initial thresholds set.

In parallel with the binary state, a continuous quality score $q_{\text{FSM}} \in [0, 1]$ is computed to facilitate downstream sensor fusion. This score linearly normalizes the measured force $F_{\text{mag}}$ between the activation threshold $F_{\text{on}}$ and the expected maximum $F_{\text{max}}$, strictly bounded to prevent invalid probabilities:
\begin{equation}
    q_{\text{FSM}} = \max\left(0, \min\left(1, \frac{F_{\text{mag}} - F_{\text{on}}}{F_{\text{max}} - F_{\text{on}}}\right)\right)
\end{equation}

\subsubsection{Detector 2: Kinematic-Based Detection (GLRT)}

Inspired by \cite{wu2025doglegs, skog2010zero}, we employ a GLRT to evaluate foot stationarity. Unlike \cite{wu2025doglegs}, this implementation relies exclusively on the main body IMU and kinematics, eliminating the need for additional foot-mounted IMUs. For a sliding window of $N_w$ velocity measurements with an isotropic covariance $\Sigma_0 = \sigma_v^2 \mathbf{I}_{3\times3}$, where $\sigma_v$ = 0.45 m/s determined empirically, we define two competing hypotheses:
\begin{itemize}
    \item \textbf{Null Hypothesis $H_0$ (Stance):} The foot is stationary. $\mathbf{v}_i \sim \mathcal{N}(\mathbf{0}, \Sigma_0)$.
    \item \textbf{Alternative Hypothesis $H_1$ (Swing):} The foot is moving. $\mathbf{v}_i \sim \mathcal{N}(\boldsymbol{\mu}, \Sigma_0)$ where $\boldsymbol{\mu} \neq \mathbf{0}$.
\end{itemize}

The standard GLRT evaluates the log-likelihood ratio of the two hypotheses: $\ln \Lambda = \ln(L_1) - \ln(L_0)$, where $L_k$ is the joint probability density function for hypothesis $H_k$. Following standard algebraic simplification for Gaussian distributions with identical covariance, the test statistic resolves to:
$$
T_{\text{standard}} = 2\ln \Lambda = N_w \hat{\boldsymbol{\mu}}^T \Sigma_0^{-1} \hat{\boldsymbol{\mu}}
$$
where $\hat{\boldsymbol{\mu}}$ is the sample mean of the velocity vectors over the data window $N_w$. Under standard statistical convention, $T_{\text{standard}}$ is evaluated against the standard 95\% confidence interval on a $\chi^2$ distribution with 3 degrees of freedom (corresponding to the velocity of the foot), which yields a critical threshold of approximately 7.815.

Since the effectiveness of data window size $N_w$ changes depending on the data rate, we normalized test statistic, $T_{\text{GLRT}}$:
$$
\frac{T_{\text{standard}}}{N_w} = T_{\text{GLRT}} = \hat{\mu}^T \Sigma_0^{-1} \hat{\mu}
$$

Consequently, the threshold must be scaled inversely. In our experiments we used $N_w = 4$ ($\approx$ 10ms), so the effective 95\% threshold becomes $\gamma_{\text{on}} \approx 7.815 / 4 \approx 2.0$. The detector outputs a discrete \textbf{STANCE} state when $T_{\text{GLRT}} < \gamma_{\text{on}}$, and \textbf{SWING} otherwise.

To facilitate sensor fusion, we map this discrete logic into a continuous quality score $q_{\text{GLRT}} \in [0, 1]$. The score equals 1.0 when the test statistic is perfectly zero, and decreases linearly as the velocity variance approaches the physical movement threshold $\gamma_{\text{on}}$:
$$
q_{\text{GLRT}} = \max\left(0, \min\left(1, 1 - \frac{T_{\text{GLRT}}}{\gamma_{\text{on}}}\right)\right)
$$

\subsubsection{Fused Score and Adaptive Covariance}

The two scores are fused using a multiplicative combination, or ``soft AND-gate":
\begin{equation}
    q_{\text{final}} = q_{\text{FSM}} \times q_{\text{GLRT}}
\end{equation}
This ensures a high final score only if the foot is \textit{both} loaded ($q_{\text{FSM}} \approx 1$) \textit{and} stationary ($q_{\text{GLRT}} \approx 1$). This final score is then used to dynamically set the $3 \times 3$ measurement noise covariance matrix, $\mathbf{R}_i$, for each foot $i$.

A high score ($q_{\text{final}} \to 1$) signifies high confidence and yields a low covariance (high trust). A low score ($q_{\text{final}} \to 0$), indicating slip, signifies low confidence and yields a high covariance, causing the ESEKF to effectively ignore the measurement. We define a baseline noise $\sigma_{\text{base}}$ (1 by default) for an ideal stance and compute the intuitive adaptive standard deviation $\sigma_i$:
\begin{equation}
\label{eqn:fusion}
    \sigma_i = \frac{\sigma_{\text{base}}}{\max(q_{\text{final}, i}, \epsilon)}
\end{equation}
where $\epsilon$ is $10^{-6}$. This constructs the final isotropic covariance matrix $\mathbf{R}_i$ that is passed to the ESEKF:
\begin{equation}
    \mathbf{R}_i = \sigma_i^2 \cdot \mathbf{I} = \left( \frac{\sigma_{\text{base}}}{\max(q_{\text{final}, i}, \epsilon)} \right)^2 \mathbf{I}
\end{equation}

\vspace{-0.2cm}

\section{EXPERIMENTS}
\label{sec:experiments}

\begin{table*}[h]
\centering
\caption{Comparison with the Proprioceptive Methods Only. Metrics are Calculated as RMSE. (\textbf{BEST}, \underline{SECOND BEST})}
\vspace{-0.3cm}
\begin{tabular}{llcccccccc}
\toprule
 & \textbf{Method} & \textbf{ATE [m]} & \textbf{AHE [deg]} & \textbf{RPE Trans [\%]} & \textbf{RPE Rot [deg/m]} & \textbf{FPE [m]} & \textbf{Drift [\%]} & \textbf{Length Err [\%]} & \textbf{Frechet [m]} \\
\midrule
\multirow{3}{*}{\rotatebox[origin=c]{90}{Concrete}} 
 & Bloesch et al. & 81.450 & 84.668 & 94.968 & 12.784 & 141.178 & 85.946 & 53.688 & 141.068 \\
 & Hartley et al. & \underline{50.785} & \underline{44.255} & \underline{68.463} & \underline{3.907} & \underline{110.346} & \underline{54.282} & \underline{12.407} & \underline{110.208} \\
 & Ours & \textbf{12.019} & \textbf{13.666} & \textbf{16.550} & \textbf{2.646} & \textbf{21.549} & \textbf{12.627} & \textbf{10.906} & \textbf{21.451} \\
\midrule
\multirow{3}{*}{\rotatebox[origin=c]{90}{Tile}} 
 & Bloesch et al. & 11.833 & \underline{30.142} & \underline{38.643} & \underline{5.600} & \underline{22.316} & \underline{20.622} & 13.355 & \underline{22.311} \\
 & Hartley et al. & \underline{11.696} & 39.793 & 51.763 & 9.451 & 26.606 & 27.032 & \underline{9.291} & 26.545 \\
 & Ours & \textbf{4.016} & \textbf{15.108} & \textbf{16.955} & \textbf{4.018} & \textbf{7.281} & \textbf{7.669} & \textbf{5.572} & \textbf{7.284} \\
\midrule
\multirow{3}{*}{\rotatebox[origin=c]{90}{Grass}} 
 & Bloesch et al. & \underline{5.315} & \underline{21.220} & \underline{33.420} & \underline{4.168} & \underline{9.609} & \underline{23.439} & 10.803 & \underline{9.574} \\
 & Hartley et al. & 6.908 & 38.565 & 47.986 & 9.684 & 15.338 & 32.791 & \underline{6.319} & 15.310 \\
 & Ours & \textbf{1.385} & \textbf{9.196} & \textbf{11.497} & \textbf{2.559} & \textbf{1.608} & \textbf{3.811} & \textbf{2.603} & \textbf{1.672} \\
\midrule
\multirow{3}{*}{\rotatebox[origin=c]{90}{Pebble}} 
 & Bloesch et al. & 5.457 & \underline{27.025} & 39.508 & \underline{3.708} & \underline{8.146} & 23.839 & 16.320 & \underline{8.132} \\
 & Hartley et al. & \underline{4.263} & 34.410 & \underline{33.666} & 8.566 & 8.500 & \underline{23.599} & \underline{3.499} & 8.703 \\
 & Ours & \textbf{1.273} & \textbf{12.133} & \textbf{14.876} & \textbf{2.823} & \textbf{1.354} & \textbf{3.749} & \textbf{3.457} & \textbf{1.437} \\
\midrule
\multirow{3}{*}{\rotatebox[origin=c]{90}{Rock}} 
 & Bloesch et al. & 30.876 & 68.094 & 102.549 & \underline{4.419} & 57.787 & 84.863 & 11.092 & 57.696 \\
 & Hartley et al. & \underline{15.028} & \underline{42.122} & \underline{62.278} & 9.994 & \underline{32.110} & \underline{47.156} & \underline{1.996} & \underline{32.075} \\
 & Ours & \textbf{6.857} & \textbf{16.042} & \textbf{33.649} & \textbf{2.427} & \textbf{10.986} & \textbf{16.135} & \textbf{1.832} & \textbf{10.979} \\
\bottomrule
\end{tabular}
\label{tab:proprioceptive}
\vspace{-0.5cm}
\end{table*}

\subsection{Experimental Setup}

\begin{figure}[h]
    \centering
    \includegraphics[width=1.0\linewidth]{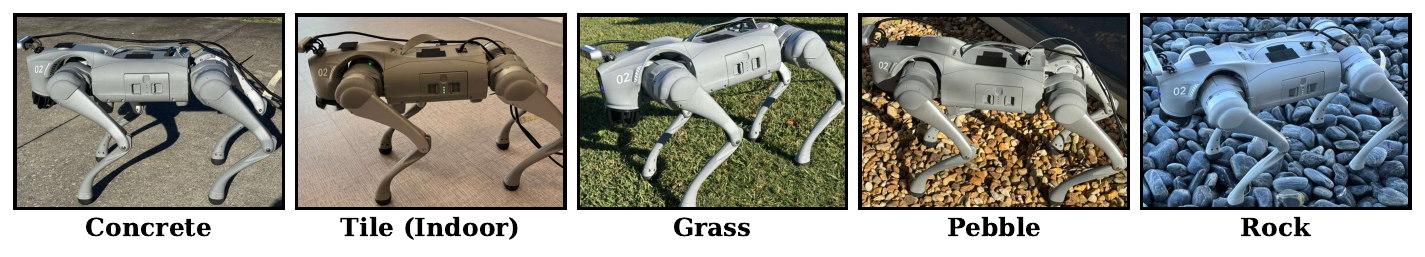}
    \caption{Our dataset spans diverse hard (concrete, tile), soft (grass), granular (pebble, rock) terrains.}
    \label{fig:terrain_images}
    \vspace{-0.6cm}
\end{figure}

To validate our proposed method, we conduct a comprehensive evaluation on our dataset (detailed in Section \ref{sec:dataset}), which spans diverse indoor and outdoor environments such as grass, pebble, and rock (Figure \ref{fig:terrain_images}). 

We evaluate the odometry accuracy of our system against seminal study in the proprioceptive leg odometry \footnote{\vspace{-0.7cm}Implementation provided by \cite{wu2025doglegs}.} \cite{bloesch2013stateVelocity} and an open-source InEKF based approach \cite{Hartley-RSS-18}. Note that both methods' parameter sets are updated according to our robot. Furthermore, we include state-of-the-art VIO methods, OpenVINS \cite{geneva2020openvins} and VINS-Fusion \cite{qin2019a}, in our comparison. Although VIO methods are not direct competitors to our proprioceptive-only approach, this comparison provides a valuable performance baseline against a system utilizing rich exteroceptive data. Our intent is to contextualize the performance achievable by proprioceptive sensing alone in environments where VIO is challenged.

We report our results on standard odometry estimation metrics, which are synchronized with ground truth via timestamps, Absolute Trajectory Error (ATE), Absolute Heading Error (AHE), and Relative Position Error (RPE). Furthermore, we also evaluate geometric error metrics including Final Position Error (FPE), total drift from the final position, trajectory length error (overshooting or undershooting the final position), and Fr\'{e}chet distance. Note that all metrics are calculated as the Root Mean Squared Error (RMSE) to strictly penalize catastrophic tracking failures.

\begin{table*}[t]
\centering
\caption{Comparison with Exteroceptive Methods (VIO). Metrics are Calculated as RMSE. (\textbf{BEST}, \underline{SECOND BEST})}
\vspace{-0.3cm}
\begin{tabular}{llcccccccc}
\toprule
 & \textbf{Method} & \textbf{ATE [m]} & \textbf{AHE [deg]} & \textbf{RPE Trans [\%]} & \textbf{RPE Rot [deg/m]} & \textbf{FPE [m]} & \textbf{Drift [\%]} & \textbf{Length Err [\%]} & \textbf{Frechet [m]} \\
\midrule
\multirow{3}{*}{\rotatebox[origin=c]{90}{Grass}} 
 & OpenVINS & 18.054 & \underline{25.747} & 88.801 & \underline{6.332} & 24.865 & 62.389 & 73.454 & 25.281 \\
 & VINS-Fusion & \underline{4.677} & 34.820 & \underline{37.002} & 8.775 & \underline{6.082} & \underline{15.568} & \underline{20.708} & \underline{6.466} \\
 & Ours & \textbf{1.385} & \textbf{9.196} & \textbf{11.497} & \textbf{2.559} & \textbf{1.608} & \textbf{3.811} & \textbf{2.603} & \textbf{1.672} \\
\midrule
\multirow{3}{*}{\rotatebox[origin=c]{90}{Pebble}} 
 & OpenVINS & 4.393 & 29.516 & 39.777 & 6.520 & 8.008 & 24.267 & 26.949 & 8.203 \\
 & VINS-Fusion & \underline{2.274} & \underline{28.145} & \underline{26.454} & \underline{6.488} & \underline{3.866} & \underline{12.384} & \underline{15.963} & \underline{3.857} \\
 & Ours & \textbf{1.273} & \textbf{12.133} & \textbf{14.876} & \textbf{2.823} & \textbf{1.354} & \textbf{3.749} & \textbf{3.457} & \textbf{1.437} \\
\midrule
\multirow{3}{*}{\rotatebox[origin=c]{90}{Rock}} 
 & OpenVINS & 27.279 & \underline{25.282} & 130.042 & \underline{5.362} & 55.647 & 81.721 & \underline{118.867} & 56.284 \\
 & VINS-Fusion & \underline{15.719} & 27.191 & \underline{118.604} & 6.926 & \underline{39.087} & \underline{57.401} & 294.253 & \underline{38.985} \\
 & Ours & \textbf{6.857} & \textbf{16.042} & \textbf{33.649} & \textbf{2.427} & \textbf{10.986} & \textbf{16.135} & \textbf{1.832} & \textbf{10.979} \\
\bottomrule
\end{tabular}
\label{tab:vio_comparison}
\end{table*}
\subsection{Quadruped Dataset}
\label{sec:dataset}
\begin{figure*}[thbp!]
    \centering
\includegraphics[width=1.0\linewidth,trim={0 0.25cm 0 0.27cm},clip]{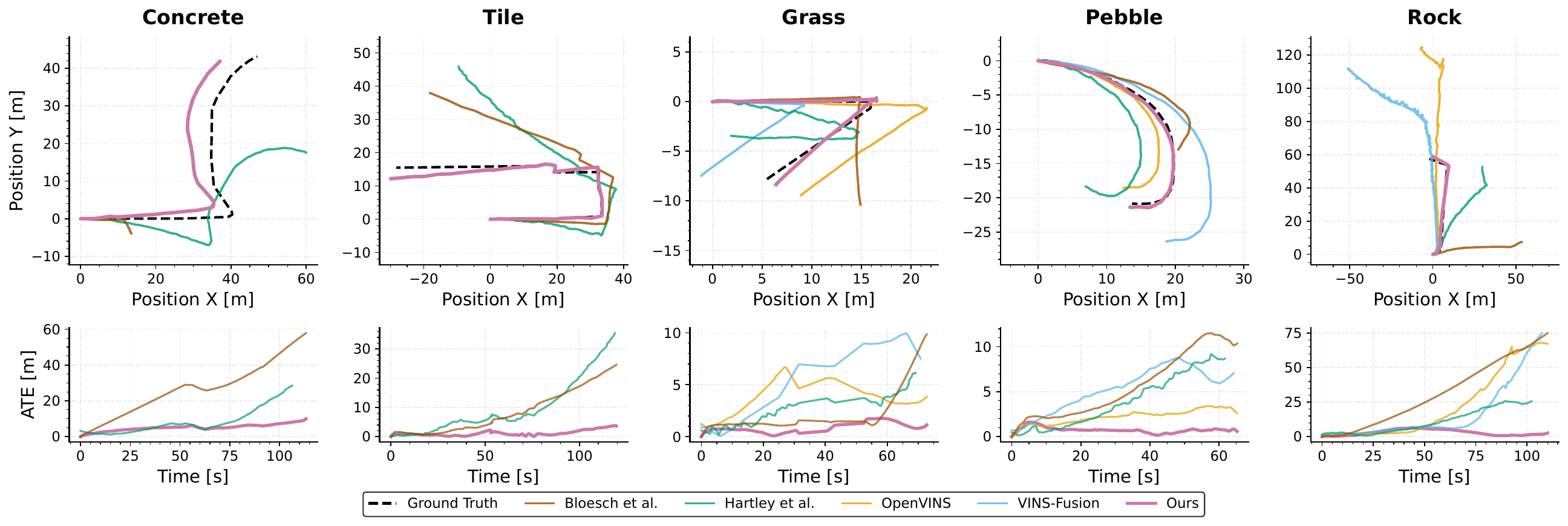}
\vspace{-0.5cm}    
    \caption{Example sequences and method's predictions are visualized (in meters) and Absolute Trajectory Error (ATE) plots over time.}
    \label{fig:seq_compare}
\vspace{-0.7cm}    
\end{figure*}
To evaluate the algorithm, we collected a challenging dataset using a Unitree Go2 robot. The sole inputs to our method are the onboard proprioceptive sensors sampled at 500Hz: IMU, joint encoders, and foot force sensors. An external depth camera (30Hz) provides synchronized RGB and point cloud data for validation only; it is not used for estimation.

\noindent We recorded 29 sequences across hard (1092m concrete, 663m tile), deformable (268m grass), and granular (235m pebble, 136m rock) terrains (Figure \ref{fig:terrain_images}). Ground truth trajectories are defined by predefined paths with surveyed waypoints and verified against synchronized video recordings. While this provides less precision than motion capture (indoor only) or RTK-GPS (outdoor), we note that: (i) the identical reference is applied uniformly across all methods to ensure fair and relative comparison without method-specific tuning, (ii) our geometric metrics (FPE, Frechet distance) assess trajectory shape independently of temporal alignment precision, and (iii) the observed performance margins (e.g., a 4–7× ATE reduction over baselines) far exceed any plausible ground truth imprecision. We also release our dataset publicly in both ROS 2 bag and post-processed formats.

\subsection{Proprioceptive Comparison}
Table \ref{tab:proprioceptive} demonstrates the performance of our method against previous proprioceptive pipelines, specifically Bloesch et al. \cite{bloesch2013stateVelocity} and Hartley et al. \cite{Hartley-RSS-18}. Our approach yields substantially lower errors across every terrain type and every metric by a wide margin.

While all three methods fundamentally rely on a variant of the Kalman Filter, the primary methodological difference lies in the measurement update formulation triggered during the leg stance phase. Bloesch et al. \cite{bloesch2013stateVelocity} utilizes a fixedthreshold on the GRF to determine stance and applies a fixed, static covariance for every measurement update, without accounting for the dynamic magnitude of the signal.

Conversely, recall that Hartley et al. \cite{Hartley-RSS-18} also adopts an adaptive contact estimation similar to ours. Their measurement covariance is dynamically scaled by the squared change in the estimated GRF: $\alpha \times (\Delta \text{GRF})^2$, where $\alpha$ is typically 10 or 100. While effective at capturing the high-frequency force spikes at initial touchdown, GRF gradually decreases as the stance phase progresses. Therefore, scaling covariance purely by $\Delta \text{GRF}$ yields suboptimal filter certainty during the stable portion of the stance cycle.

Although this measurement model allows Hartley et al. \cite{Hartley-RSS-18} to estimate total traveled distance (scale) better than Bloesch et al. \cite{bloesch2013stateVelocity}, it does not fully correct the underlying yaw divergence. Figure \ref{fig:seq_compare} visualizes the estimated trajectories on example sequences. The measurement model of Hartley et al. struggles to maintain heading accuracy across all sequences. On the other hand, Bloesch et al. \cite{bloesch2013stateVelocity} exhibits scale drift on concrete and pebble surfaces, while simultaneously exhibiting heading estimation errors on all terrain types.

\subsection{Exteroceptive Comparison}
\vspace*{-0.1cm}
Although our proposed method relies exclusively on proprioceptive sensors, we benchmark its performance against established monocular VIO systems, specifically OpenVINS \cite{geneva2020openvins} and VINS-Fusion \cite{qin2019a}. While these VIO pipelines utilize graph-based smoothing rather than filter-based estimation, they inherently propagate the system state via the IMU and apply visual feature tracking as measurement updates, directly analogous to our contact-driven zero-velocity updates. We conducted camera-IMU extrinsic and time sync offset calibration using Kalibr tool \cite{rehder2016extending}. 

Table \ref{tab:vio_comparison} summarizes the quantitative performance of our method and the baseline VIO models on grass, pebble, and rock terrains, with corresponding qualitative example trajectories visualized in Figure \ref{fig:seq_compare}. As demonstrated, our proprioceptive approach outperforms both VIO systems, achieving this without the severe computational overhead required to process high-dimensional feature-rich RGB images.

Note that, concrete and tile sequences are excluded from this comparison. These sequences were recorded in highly dynamic environments with reflective surfaces lacking distinct features. The lack of stable visual landmarks induces severe feature tracking failures, inevitably leading to the divergence of the camera pose optimization \cite{thalhammer2024challenges}. Despite our best efforts to tune, neither system completed these trajectories with geometrically reasonable outputs.


\subsection{Effect of FSM and GLRT}

In this ablation study, we evaluate the independent effectiveness of the FSM-based and GLRT-based contact detectors. A stable stance requires both significant ground reaction force and low kinematic velocity. The isolated GMM-FSM accurately identifies mechanically loaded contacts, but relying purely on force data leaves the system susceptible to unmeasured kinematic slippage. Conversely, the GLRT evaluates kinematic foot stationarity but remains completely blind to physical ground reaction forces, making it inherently susceptible to noise amplified by the forward kinematics calculations.
\begin{figure*}[thbp]
    \centering
\includegraphics[width=0.95\linewidth,trim={0 0.27cm 0 0.27cm},clip]{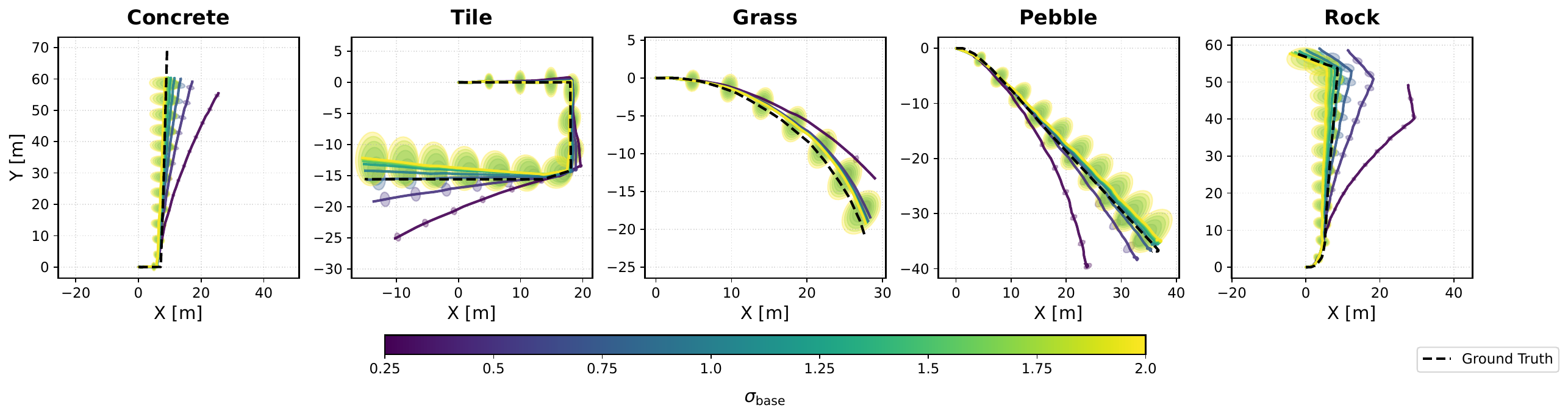}
\vspace{-0.3cm}    
    \caption{The ability to correct drift caused by IMU is governed by the measurement covariance even under identical contact conditions.}
    \label{fig:covariance_seqs}
\vspace{-0.5cm}    
\end{figure*}


Table \ref{tab:ablation_standard} presents the quantitative performance of the isolated FSM and GLRT detectors against the fused model. The central finding is not that fusion always achieves the lowest ATE, but that each individual detector exhibits terrain dependent failure modes that fusion eliminates. The GLRT-only detector accumulates 23.161m ATE on concrete and nearly double the fused model (12.025m) since forward kinematic noise accumulates over long trajectories (up to 300 m). In contrast, the FSM only detector fails on tile sequence~1 (ATE 9.577m Table~\ref{tab:indoor_standard}), where force based detection alone misclassifies contact events. The fused model recovers this failure (ATE 3.730m). On grass, all configuration perform within 0.24m which is a margin negligible relative to the observed error magnitude. On the challenging granular terrains (i.e., pebble and rock), where slip detection is more critical, the fused model achieves the lowest ATE in both cases (1.272m and 6.862m). This aligns with our theoretical motivation, a foot must be both physically loaded (FSM) and kinematically stationary (GLRT) to produce a reliable measurement. The fused model sacrifices marginal optimality on benign terrain for consistent robustness across the full terrain spectrum. This provides a useful property for real world deployment where terrain type is unknown \textit{a priori}.

\vspace{-0.2cm}
\subsection{Covariance Quantification}

Finally, we evaluate the impact of measurement covariance quantification on overall trajectory estimation. A sensitivity analysis is performed by modulating the estimated covariance matrix with a scalar coefficient $\sigma_{\text{base}}$ (Eqn. \ref{eqn:fusion}) under identical contact conditions. The proposed baseline formulation utilizes a default coefficient of $\sigma_{\text{base}} = 1$.  By systematically varying this scalar, we observe the magnitude of contact uncertainty dictates yaw estimation accuracy; improper uncertainty inevitably induces severe lateral drift. As demonstrated in Figure \ref{fig:covariance_seqs}, the baseline approach ($\sigma_{\text{base}} = 1$) may marginally overestimate or underestimate the true physical covariance depending on the specific terrain sequence. 

Specifically, on rocky terrain, artificially setting $\sigma_{\text{base}} = 0.25$ forces a significant trajectory divergence due to underestimation of the true measurement covariance. Crucially, this specific divergence pattern is geometrically similar to the trajectory estimated by Hartley et al. (Figure \ref{fig:seq_compare}) on the corresponding rocky terrain. This visual and mathematical equivalence aligns with our proposition that the covariance formulation adopted by Hartley et al. systematically underestimates the true covariance during the stable stance phase. Because their approach scales covariance inversely with the change in ground reaction force ($\Delta \text{GRF}$), it is overly sensitive to the initial high-frequency touchdown spikes and subsequently assigns an erroneously low covariance to the remainder of the stance cycle. 

Ultimately, accurately quantifying the measurement update covariance remains a complex algorithmic challenge, as it necessitates the direct mathematical transformation of an abstract binary classification uncertainty into a physically bounded, real-world covariance matrix.

\begin{table}[t!]
\centering
\caption{Effect of Different Contact Models on Standard Metrics (\textbf{BEST}, \underline{SECOND BEST}).}
\vspace{-0.2cm}
\begin{tabular}{llcccc}
\toprule
 & \textbf{Experiment} & \begin{tabular}{@{}c@{}}\textbf{ATE} \\ \textbf{[m]}\end{tabular} & \begin{tabular}{@{}c@{}}\textbf{AHE} \\ \textbf{[deg]}\end{tabular} & \begin{tabular}{@{}c@{}}\textbf{RPE Trans} \\ \textbf{[\%]}\end{tabular} & \begin{tabular}{@{}c@{}}\textbf{RPE Rot} \\ \textbf{[deg/m]}\end{tabular} \\
\midrule
\multirow{3}{*}{\rotatebox[origin=c]{90}{Concrete}} 
 & FSM Only & \textbf{11.976} & \textbf{13.458} & \underline{16.675} & \textbf{2.584} \\
 & GLRT Only & 23.161 & 19.343 & 29.111 & 3.267 \\
 & FSM + GLRT & \underline{12.025} & \underline{13.668} & \textbf{16.555} & \underline{2.646} \\
\midrule
\multirow{3}{*}{\rotatebox[origin=c]{90}{Tile}} 
 & FSM Only & 4.775 & 16.235 & 20.994 & \underline{3.815} \\
 & GLRT Only & \textbf{1.614} & \textbf{12.476} & \textbf{13.642} & \textbf{3.551} \\
 & FSM + GLRT & \underline{4.018} & \underline{15.108} & \underline{16.956} & 4.018 \\
\midrule
\multirow{3}{*}{\rotatebox[origin=c]{90}{Grass}} 
 & FSM Only & \textbf{1.169} & \underline{9.734} & 12.207 & 2.678 \\
 & GLRT Only & 1.413 & 10.259 & \underline{11.809} & \underline{2.613} \\
 & FSM + GLRT & \underline{1.388} & \textbf{9.199} & \textbf{11.503} & \textbf{2.559} \\
\midrule
\multirow{3}{*}{\rotatebox[origin=c]{90}{Pebble}} 
 & FSM Only & 1.372 & 12.372 & 15.133 & 2.832 \\
 & GLRT Only & \underline{1.311} & \textbf{11.735} & \textbf{13.660} & \textbf{2.815} \\
 & FSM + GLRT & \textbf{1.272} & \underline{12.135} & \underline{14.880} & \underline{2.823} \\
\midrule
\multirow{3}{*}{\rotatebox[origin=c]{90}{Rock}} 
 & FSM Only & 7.729 & 18.167 & 37.212 & 2.456 \\
 & GLRT Only & \underline{7.141} & \underline{16.933} & \underline{35.040} & \underline{2.427} \\
 & FSM + GLRT & \textbf{6.862} & \textbf{16.050} & \textbf{33.662} & \textbf{2.427} \\
\bottomrule
\end{tabular}
\label{tab:ablation_standard}
\end{table}
\vspace{-0.3cm}
\begin{table}[t!]
\centering
\caption{Effect of Different Contact Models in Tile on Standard Metrics (\textbf{BEST}, \underline{SECOND BEST}).}
\vspace{-0.3cm}
\begin{tabular}{llcccc}
\toprule
\textbf{\#} & \textbf{Experiment} & \begin{tabular}{@{}c@{}}\textbf{ATE} \\ \textbf{[m]}\end{tabular} & \begin{tabular}{@{}c@{}}\textbf{AHE} \\ \textbf{[deg]}\end{tabular} & \begin{tabular}{@{}c@{}}\textbf{RPE Trans} \\ \textbf{[\%]}\end{tabular} & \begin{tabular}{@{}c@{}}\textbf{RPE Rot} \\ \textbf{[deg/m]}\end{tabular} \\
\midrule
\multirow{3}{*}{\rotatebox[origin=c]{0}{\textbf{1}}} 
 & FSM Only & 9.577 & 29.044 & 46.048 & \textbf{3.441} \\
 & GLRT Only & \textbf{3.252} & \underline{18.633} & \underline{20.284} & 4.230 \\
 & FSM + GLRT & \underline{3.730} & \textbf{17.578} & \textbf{19.716} & \underline{3.974} \\
\midrule
\multirow{3}{*}{\rotatebox[origin=c]{0}{2}} 
 & FSM Only & 1.756 & \textbf{20.095} & 14.330 & \textbf{7.394} \\
 & GLRT Only & \textbf{1.247} & 21.216 & \textbf{12.574} & 7.758 \\
 & FSM + GLRT & \underline{1.562} & \underline{20.778} & \underline{13.246} & \underline{7.440} \\
\midrule
\multirow{3}{*}{\rotatebox[origin=c]{0}{3}} 
 & FSM Only & \textbf{1.007} & \textbf{5.020} & \textbf{6.372} & \textbf{1.206} \\
 & GLRT Only & 1.854 & 6.984 & 7.706 & \underline{1.676} \\
 & FSM + GLRT & \underline{1.811} & \underline{6.937} & \underline{7.336} & 1.696 \\
\midrule
\multirow{3}{*}{\rotatebox[origin=c]{0}{4}} 
 & FSM Only & \underline{0.645} & \textbf{5.522} & 10.721 & \textbf{1.510} \\
 & GLRT Only & \textbf{0.557} & 6.659 & \textbf{8.904} & \underline{1.735} \\
 & FSM + GLRT & 0.776 & \underline{6.559} & \underline{9.548} & 1.807 \\
\midrule
\multirow{3}{*}{\rotatebox[origin=c]{0}{5}} 
 & FSM Only & 1.616 & \underline{7.381} & 19.542 & \underline{2.022} \\
 & GLRT Only & \textbf{1.161} & 8.890 & \underline{18.743} & 2.356 \\
 & FSM + GLRT & \underline{1.211} & \textbf{7.026} & \textbf{18.659} & \textbf{1.944} \\
\bottomrule
\end{tabular}
\label{tab:indoor_standard}
\vspace{-0.4cm}
\end{table}

\section{CONCLUSION}

\addtolength{\textheight}{-0.0cm}   








In this paper, we presented a proprioceptive ESEKF framework for robust quadruped robot odometry. Our architectural advantage of novel fusion of adaptive GMM-FSM and kinematic GLRT contact detection algorithms ensures stable state estimation across rigid, compliant, and granular terrains. 

Experiments conclusively demonstrate that our proposed methodology significantly reduces trajectory drift compared to existing proprioceptive baselines. The ablation studies prove the critical importance of accurately quantifying measurement update covariance to prevent yaw divergence. To facilitate reproducibility and further research, we release our challenging multi-terrain quadruped dataset alongside the complete open-source ROS 2 implementation of our algorithmic pipeline. 

Future work will focus on integrating our adaptive, dual-modality contact estimation into graph-based smoothing frameworks. Furthermore, we will explore the direct fusion of this robust proprioceptive backbone with exteroceptive modalities, specifically monocular VIO, to mitigate feature-tracking failures in dynamic environments.

\vspace{-0.5cm}

\bibliographystyle{unsrt}
\bibliography{myreferences}

\end{document}